\documentclass{article}
\usepackage{ijcai26}

\usepackage{times}
\usepackage{soul}
\usepackage[hyphens]{url}
\usepackage[hidelinks,breaklinks]{hyperref}
\IfFileExists{xurl.sty}{\usepackage{xurl}}{}
\usepackage[utf8]{inputenc}
\usepackage[small]{caption}
\usepackage{graphicx}
\usepackage{amsmath}
\usepackage{amsthm}
\usepackage[switch]{lineno}

\AtBeginDocument{}
\usepackage{xcolor}
\usepackage{tabularx}
\usepackage{algorithm}
\usepackage{algorithmic}
\newcommand{\Require}{\REQUIRE}
\newcommand{\Ensure}{\ENSURE}
\newcommand{\State}{\STATE}
\newcommand{\While}[1]{\WHILE{#1}}
\newcommand{\EndWhile}{\ENDWHILE}
\newcommand{\If}[1]{\IF{#1}}
\newcommand{\Else}{\ELSE}
\newcommand{\EndIf}{\ENDIF}

\usepackage{booktabs}
\usepackage{multirow}
\usepackage{array}
\IfFileExists{makecell.sty}{\usepackage{makecell}}{}

\definecolor{commentcolor}{RGB}{0, 128, 128}

\newlength{\KADEXCompactGap}
\makeatletter
\renewcommand\paragraph{\@startsection{paragraph}{4}{\z@}{0pt}{-1em}{\normalsize\bf}}
\makeatother

\title{Sustainable Intelligence for the Wild: Democratizing Ecological Monitoring via Knowledge-Adaptive Edge Expert Agents}

\author{
Jiaxing Li$^1$
\and
Hao Fang$^1$\and
Chi Xu$^{1}$\and
Miao Zhang$^1$\and
Jiangchuan Liu$^1$\and
William	I. Atlas$^{2}$\and \\
Katrina M. Connors$^3$\And
Mark A. Spoljaric$^4$\\
\affiliations
$^1$Simon Fraser University, Vancouver, Canada\\
$^2$Wild Salmon Center, Portland, USA\\
$^3$Pacific Salmon Foundation, Vancouver, Canada\\
$^4$Haida Fisheries Program, Skidegate, Canada\\
\emails
\{jla641,fanghaof,chix,mza94,jcliu\}@sfu.ca,
watlas@wildsalmoncenter.org,\\
kconnors@psf.ca,
mark.spoljaric@haidanation.com
}

\begin{document}

\maketitle

\begin{abstract}

Rapid biodiversity loss underscore the urgency of effective monitoring, yet manual surveys remain resource-intensive. While on-device AI offers a scalable alternative, its performance in the wild is often challenged by environmental variability. Current methods rely heavily on cloud resource, which requires continuous uploading of field data for model retraining. This approach is unsuitable for remote deployments because it consumes limited power and network connectivity. To address these constraints, this research proposes a shift from model adaptation to knowledge adaptation. We introduce an architecture that separates visual perception from reasoning, combining a visual encoder with a dynamic knowledge base. We uses an explicit knowledge base to replace implicitly encoding expert knowledge into model parameters. This method also supports knowledge sustainability by preserving expert insights in a structured form. Through cross-disciplinary collaboration with biologists and Indigenous communities, this work advances ethical AI co-development, fostering responsible and culturally informed ecosystem management.

\end{abstract}

\section{Problem Statement}
\begin{figure}[t]
\centering
\includegraphics[width=0.9\linewidth]{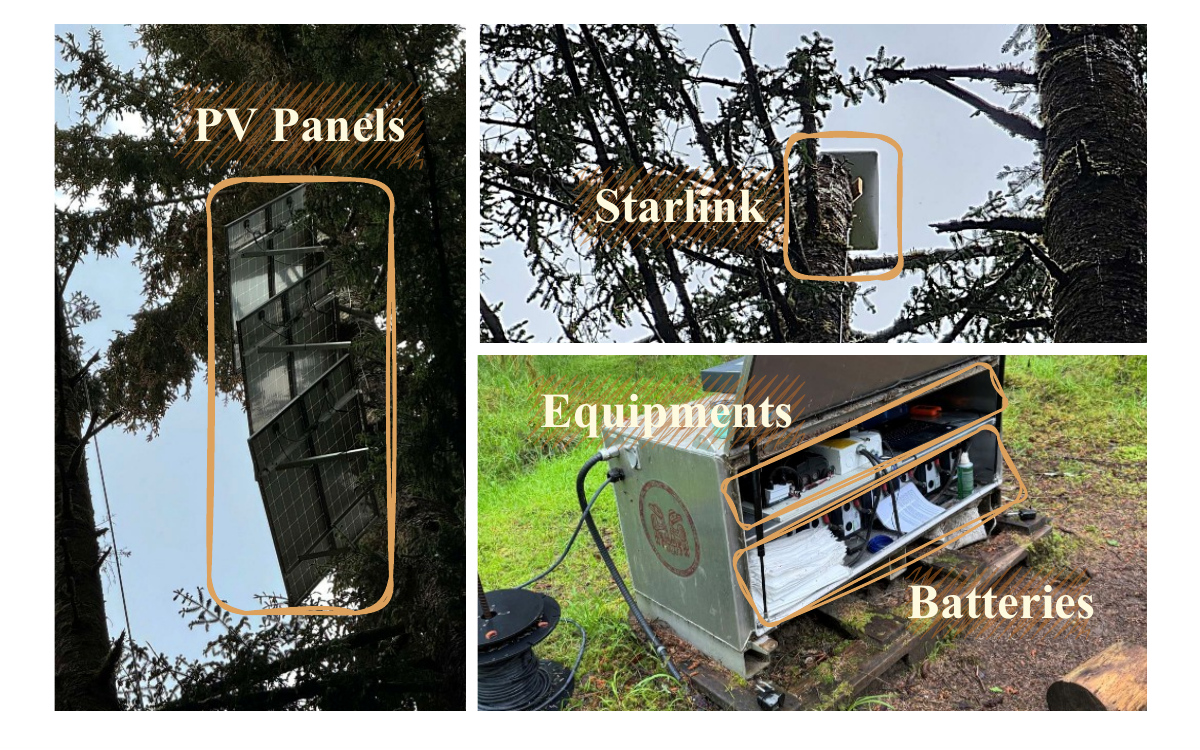} \vspace{-0.25cm}
\caption{The Off-grid Reality: Our pilot station relies entirely on solar harvesting (PV Panels) and satellite transmission (Starlink), creating strict energy budgets.} \label{fig:Hardware}
\end{figure}

\begin{figure*}[t]
\centering
\includegraphics[width=1\linewidth]{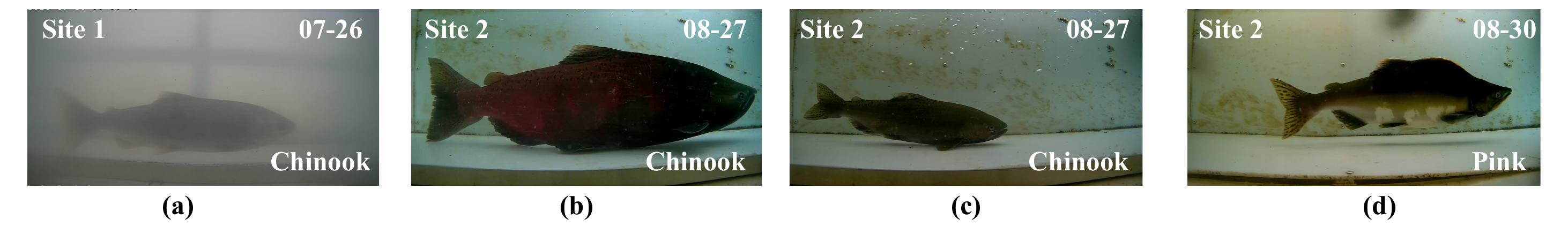}
\vspace{-0.54cm}
\caption{The Adaptation Gap: Visual traits of salmon species drift significantly across space (Estuary vs. Spawning grounds) and time, causing static models to fail.} \label{fig:Drift}
\end{figure*}

Global ecosystems are undergoing rapid disruption, driving widespread species declines across terrestrial, marine, and freshwater habitats, with freshwater species experiencing the steepest losses~\cite{wwf2024living,waples2008evolutionary,di2016multi,frolicher2018emerging,kilduff2015changing,dorner2018spatial}. Reliable wildlife sensing and analysis is therefore essential for timely, evidence-based conservation decisions~\cite{kshitiz23bird,gordon23rhinos,musmanni2023protecting,atlas2021indigenous,schindler2015prediction}. Yet in practice, ecological monitoring still scales through manual surveys and expert annotation: modern camera-trap and in-river video systems can generate millions of images per season, while expert review requires tens of thousands of hours—creating persistent data backlogs and delayed management actions~\cite{naturetech2025ground,norouzzadeh2018automatically}.

Prior work has progressively shifted salmon monitoring from manual review to automated sensing and learning. Video-based weir systems first applied computer vision to expedite in-season fish counting~\cite{atlas2023wild}, while sonar-based pipelines addressed low-visibility rivers through detection and tracking in sonar videos~\cite{kay2022caltech,kay2024align}. SALINA further advanced this direction with real-time sonar analytics and energy-aware deployment in remote ecosystems within Indigenous territories~\cite{xu2024salina}. \cite{xu2025exploring} explored multimodal foundation AI with expert-in-the-loop workflows to support species identification, counting, and length measurement for sustainable fisheries management in infrastructure-limited Indigenous rivers. FUSED further studied sovereignty-preserving multimodal retrieval across independently governed ecological data domains~\cite{xu2026fused}. Yet field deployment remains dominated by the prevailing \textbf{model adaptation} paradigm (i.e. continually uploading raw data for annotation and cloud retraining, then downloading and deploying updated model weights.)~\cite{xu2025exploring,folkman2025data}, which still breaks under off-grid operation and fails to preserve expert logic. This gap leads to two core barriers: an operational barrier and an knowledge  barrier.

\subsection{The Operational Barrier: Infrastructure Constrained AI in Off-Grid Rivers}

The first barrier is operational. Many priority watersheds are \emph{remote ecosystems that lack basic infrastructure support}, where deployments must sustain sensing, compute, and data delivery under strict resource constraints~\cite{axford2024collectively}. In our pilot station (Figure~\ref{fig:Hardware}), the entire pipeline runs off-grid using solar harvesting (PV panels) and battery storage, while wide-area communication depends on a high-wattage satellite terminal (Starlink). This coupling creates a hard trade-off between \emph{ecological coverage} and \emph{system uptime}: continuous connectivity competes directly with the energy budget required for sensing, local storage, and on-device inference~\cite{xu2024salina}.

Our pilot deployment directly reveals the severity of this constraint. In temperate rainforest conditions with frequent overcast periods, maintaining always-on connectivity quickly depletes battery reserves to safety cutoffs (30\%) within 72 hours, forcing a departure from ``cloud-first'' assumptions toward sparse, low-power communication that restricts connectivity to short daily windows ($\sim$2 hours/day).

These realities further expose the limits of the prevailing \textbf{model adaptation} paradigm~\cite{xu2025exploring,folkman2025data}. In infrastructure-constrained wild, model updates often require repeatedly transferring gigabytes of raw video over satellite links (e.g., Starlink with an average uplink of 14.84 Mbps~\cite{hill2025starlink}) for cloud annotation and retraining, and then pushing updated checkpoints back to the field. This closed loop is costly in energy, bandwidth, and turnaround time, and increase operational cost and carbon footprint~\cite{yang2024does,xu2024salina}.

\subsection{The Knowledge Barrier: From Vanishing Expertise to Executable Digital Heritage}

The second barrier is knowledge. Salmon appearance and environmental context shift rapidly across space and time (Figure~\ref{fig:Drift}). Even within a single species, visual traits drift throughout the migration cycle: a Chinook salmon near an estuary appear silvery and streamlined, it transforms into darker or bright red spawning forms with distinct skeletal changes as it moves upstream~\cite{axford2024collectively}.

The model adaptation paradigm attempts to absorb field expertise by training on large-scale datasets, implicitly encoding domain knowledge into model parameters. However, spatiotemporal drift can still cause performance degradation when models are deployed across different sites, seasons, and life stages. Meanwhile, failures are hard to trace, and new knowledge cannot be iterated or incorporated quickly in a principled, transparent way~\cite{xu2025exploring,liu24fish}.

Furthermore, our stakeholder engagements underscore a compounding crisis: expert scarcity and the challenge of retaining practical field expertise~\cite{wu2022survey}. High-quality monitoring relies heavily on senior field biologists and Indigenous stewardship practitioners, whose knowledge is often tacit, highly localized, and recorded only in fragmented notes~\cite{paul2024simple,ma2024leo,liu24fish}. As experts retire or rotate across watersheds, this ``river logic'' risks being lost—precisely when climate-driven variability increases the need for adaptive decisions.

This motivates a shift beyond black-box model adaptation and single-metric ``better performance'': through active collaboration with domain experts, we pursue a new application paradigm that jointly targets (i) \textbf{sustainable AI} for infrastructure-constrained wild deployments and (ii) \textbf{knowledge sustainability} that preserves and operationalizes localized stewardship logic.

\section{Alignment with Sustainable Development Goals and the LNOB principle}
The 2030 Agenda for Sustainable Development outlines a blueprint for global peace and prosperity~\cite{un-2030agenda} . A core tenet, the Leaving No One Behind (LNOB) principle, underscores the necessity of extending technological advancements to vulnerable communities and fragile ecosystems~\cite{un-lnob}. This research shifts from data-intensive cloud retraining to bandwidth-efficient knowledge adaptation. In doing so, our architecture aligns with several Sustainable Development Goals (SDGs):

\paragraph{SDG 13: Climate Action (Green AI).} Conventional AI monitoring incurs high carbon costs due to frequent cloud retraining and high-bandwidth transmission. This research supports SDG 13.2 (``Integrate climate change measures'') by adopting a resource-efficient architecture. The system minimizes computational energy by substituting intensive parameter updates with lightweight knowledge graph maintenance. Additionally, it reduces communication costs by transmitting compact semantic text instead of raw video.

\paragraph{SDG 14: Life Below Water \& SDG 15: Life on Land.} Effective biodiversity conservation depends on timely, granular ecological data, yet manual surveys remain prohibitively expensive and difficult to scale. This research advances SDG 14.2 (``Sustainably manage marine ecosystems'') and SDG 15.5 (``Halt biodiversity loss'') by deploying autonomous agents capable of fine-grained species identification and behavioral analysis in the wild.

\paragraph{SDG 17: Partnerships for the Goals.} Sustainable conservation depends on integrating computer science with ecology. We establish a unified architecture that connects AI researchers, field biologists, and conservation agencies to facilitate the transfer of technology into ecological practice. This research advances SDG 17.16 (``Enhance the Global Partnership'') and SDG 17.17 (``Encourage multi-stakeholder collaborations'') by mobilizing expertise across disciplines.

\paragraph{Alignment with LNOB.} A ``digital divide'' currently restricts the application of advanced AI in remote conservation areas. This research aligns with LNOB by democratizing and preserving expert knowledge. The system formalizes tacit expertise into updateable knowledge graphs. This process safeguards critical scientific insights that are at risk of being lost. Ultimately, the project empowers local communities with professional-grade monitoring tools,

\section{Strategies}
\begin{figure}[t]
\centering
\includegraphics[width=1\linewidth]{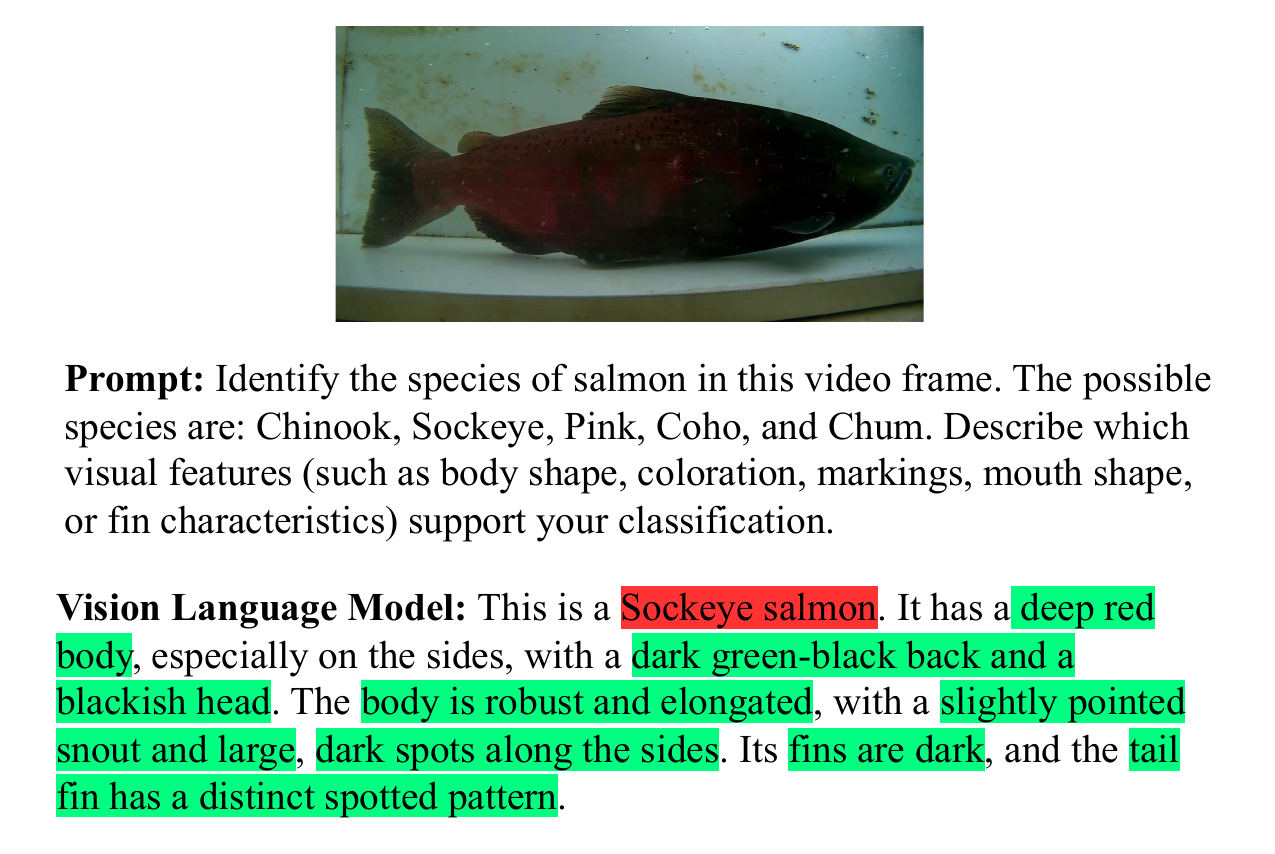}
\vspace{-0.7cm}
\caption{The semantic gap: A general-purpose VLM (Qwen3-VL-8B) identifies visual cues (green) but hallucinates the species label (red) without domain-specific exclusionary logic.}
\label{fig:VLM_Fail}
\end{figure}

To overcome the operational and knowledge barriers identified above, we propose \textbf{Knowledge-Adaptive Edge Expert Agents (KADEX)} as a strategic pivot from \textbf{model adaptation} to \textbf{knowledge adaptation}. Today’s wild on-site counting pipelines are still dominated by pure vision detectors (e.g.,  YOLO~\cite{wang2024yolov10} and RT-DETR~\cite{zhao2024detrs}), while recent attempts to replace them with VLMs remain constrained by the same two field realities: limited edge compute only sustains small-parameter models, and end-to-end VLM retraining/fine-tuning amplifies data, energy, and bandwidth costs. KADEX is motivated by a key observation: for small parameter VLMs, the dominant bottleneck is often missing expert logic, and this logic does not need to be permanently embedded in model weights. Instead, it can be externalized, updated, and retrieved on demand as a lightweight knowledge layer.

\begin{figure}[t]
\centering
\includegraphics[width=1\linewidth]{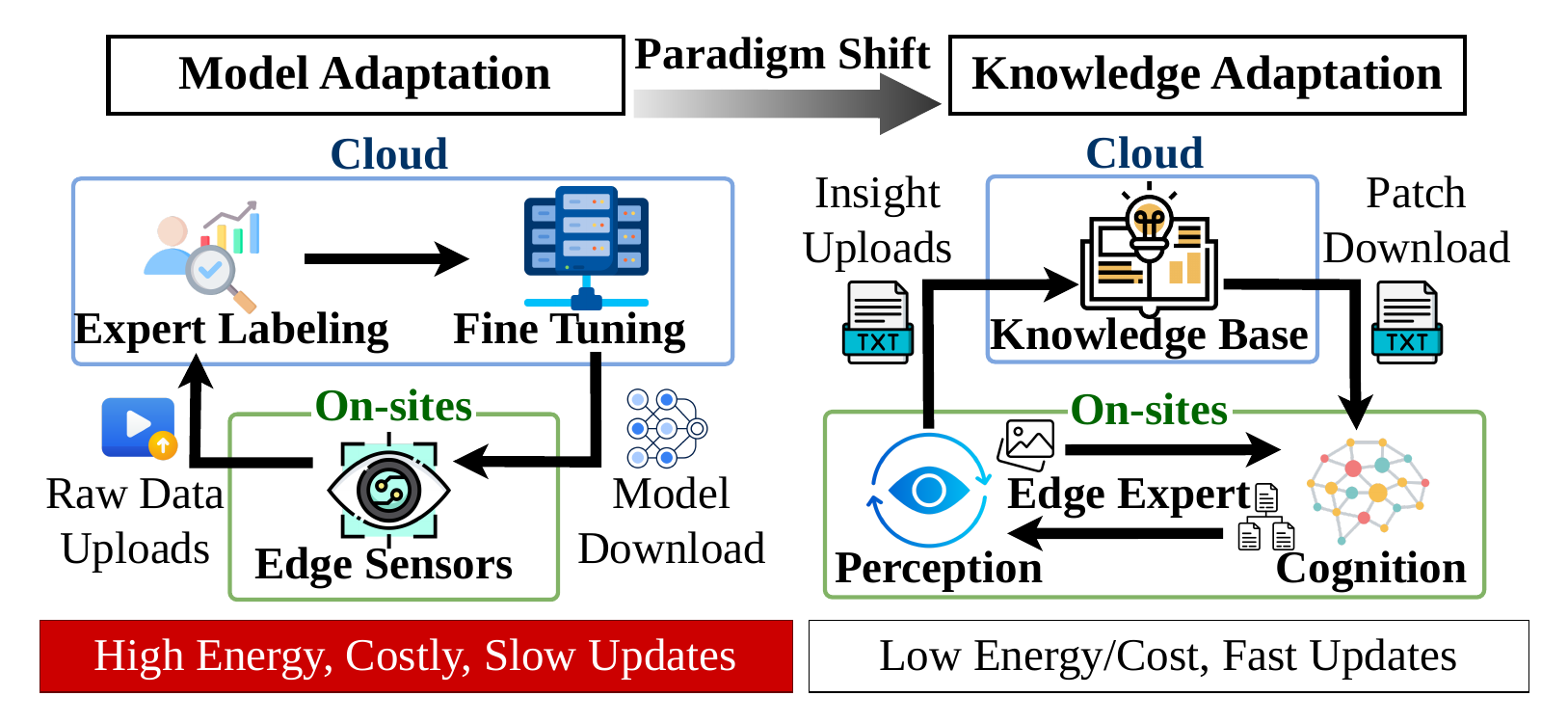}
\vspace{-0.43cm}
\caption{Strategic paradigm shift: Instead of retraining heavy models in the cloud (high energy/cost), KADEX updates a lightweight external knowledge base on the edge (low energy/cost).}
\label{fig:overview}
\end{figure}

\begin{figure*}[t]
\centering
\includegraphics[width=0.95\linewidth]{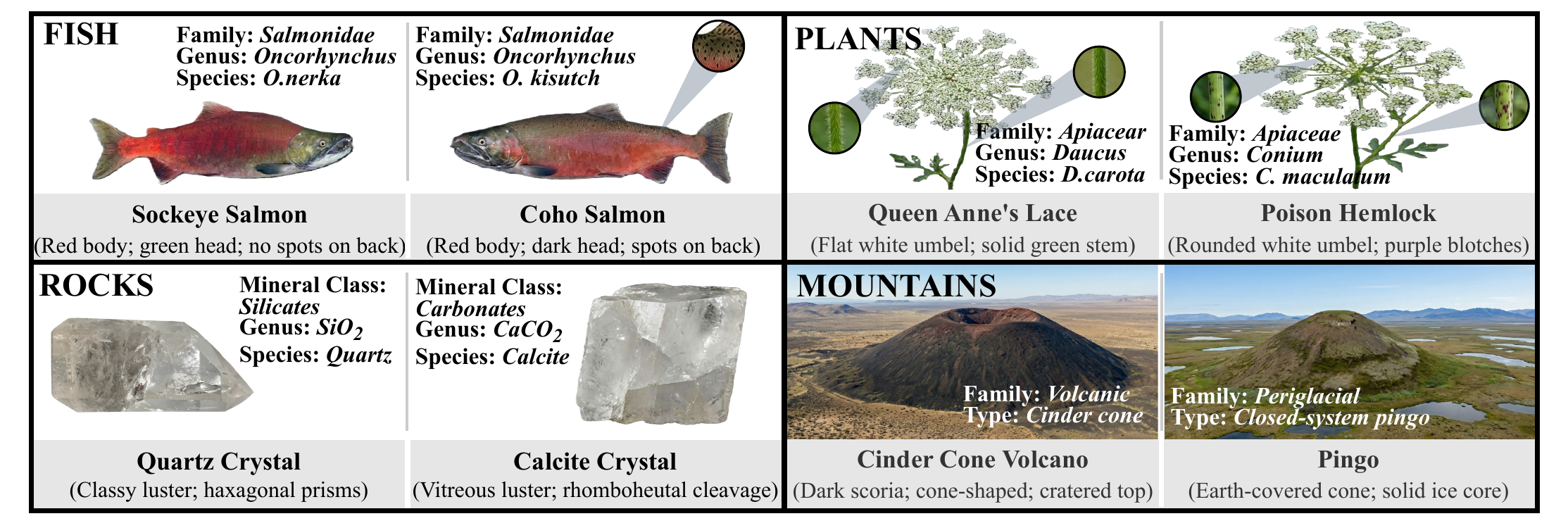}
\vspace{-0.1cm}
\caption{Visual ambiguity vs.\ taxonomic precision: Distinguishing visually similar but scientifically distinct entities extends beyond salmon to plants, minerals, and landforms.}
\label{fig:doppelgangers}
\end{figure*}

\subsection{Knowledge Adaptation to Compensate VLMs}

Figure~\ref{fig:VLM_Fail} illustrates the opportunity behind KADEX. Even when a small parameter VLM (Qwen3-VL-8B~\cite{bai2025qwen3}) fails at fine-grained species identification, its visual encoder can still extract discriminative field features (e.g., spots, body color, shape). The failure arises when these cues are mapped to labels without the exclusionary and contextual rules that experts apply (e.g., ``Sockeye do not have back/tail spots''). KADEX decouples expert logic from model parameters to avoid expensive retraining. It achieves this by using retrieval-augmented generation (RAG)~\cite{lewis2020rag,li2024structrag,yu2024visrag} to inject local rules and constraints into a lightweight knowledge base. This enables rapid, low-bandwidth iteration of domain knowledge while keeping on-device inference feasible.

\subsection{Decoupled Perception and a Transferable Knowledge Layer}

Operationally, KADEX implements this pivot by decoupling perception from cognition. A stable, lightweight perception module (e.g. CLIP~\cite{radford2021learning}) extracts visual evidence, while a dynamic retrieval-based knowledge base provides evolving expert constraints. Figure~\ref{fig:overview} shows that adaptation is driven by updating the external knowledge layer, rather than repeatedly retraining and redeploying model weights. This design keeps iteration feasible under sparse connectivity and tight energy budgets.

More broadly, KADEX targets a recurring challenge in ecological recognition: visual cues alone are often insufficient for taxonomic decisions, even with strong encoders. Figure~\ref{fig:doppelgangers} illustrates this ``lookalike'' regime, where visually similar entities require scientifically precise distinctions. KADEX couples visual evidence with explicit, updateable constraints and spatiotemporal priors, enabling transfer beyond salmon to other biodiversity and environmental sensing tasks where domain logic is essential.

Two simple cases illustrate this logic beyond salmon. For visual ambiguity, \textit{Chromis viridis} and \textit{C. atripectoralis} are near-identical blue-green reef fishes in the Indo-Pacific. Experts separate them by the black pectoral-axil mark. KADEX represents this as an attribute-level rule: an activated ``black pectoral-axil spot'' node supports \textit{C. atripectoralis} and conflicts with \textit{C. viridis}. For spatial and part-based ambiguity, consider a camera trap near Vancouver, BC, Canada. A partially occluded small felid with a short tail may match both \textit{Lynx rufus} and \textit{Lynx canadensis} from visual evidence alone. This case is useful because \textit{L. canadensis} is rarely expected in the Lower Mainland, while \textit{L. rufus} is locally plausible in southern coastal B.C. KADEX activates context nodes and locally relevant species communities in the L-SKG. This can favor \textit{Lynx rufus} and produce an auditable explanation. In both cases, expert notes become reusable structured patches rather than new model weights.

\section{Methods}
To resolve the operational conflict between strict resource limits and the need for continuous adaptation, we propose KADEX. This architecture shifts the adaptation paradigm from heavy model retraining to lightweight knowledge management. KADEX separates perception from cognition. It employs a fixed vision encoder to extract features and a Local Structured Knowledge Graph (L-SKG) to interpret them via compact updates. However, this design introduces three specific implementation challenges in off-grid scenarios: \emph{(i)} identifying ambiguous observations that require expert review; \emph{(ii)} scheduling data exchange under fluctuating energy budgets; and \emph{(iii)} maintaining a compact knowledge base within finite storage.

\begin{figure}[t]
    \centering
    \includegraphics[width=1\linewidth]{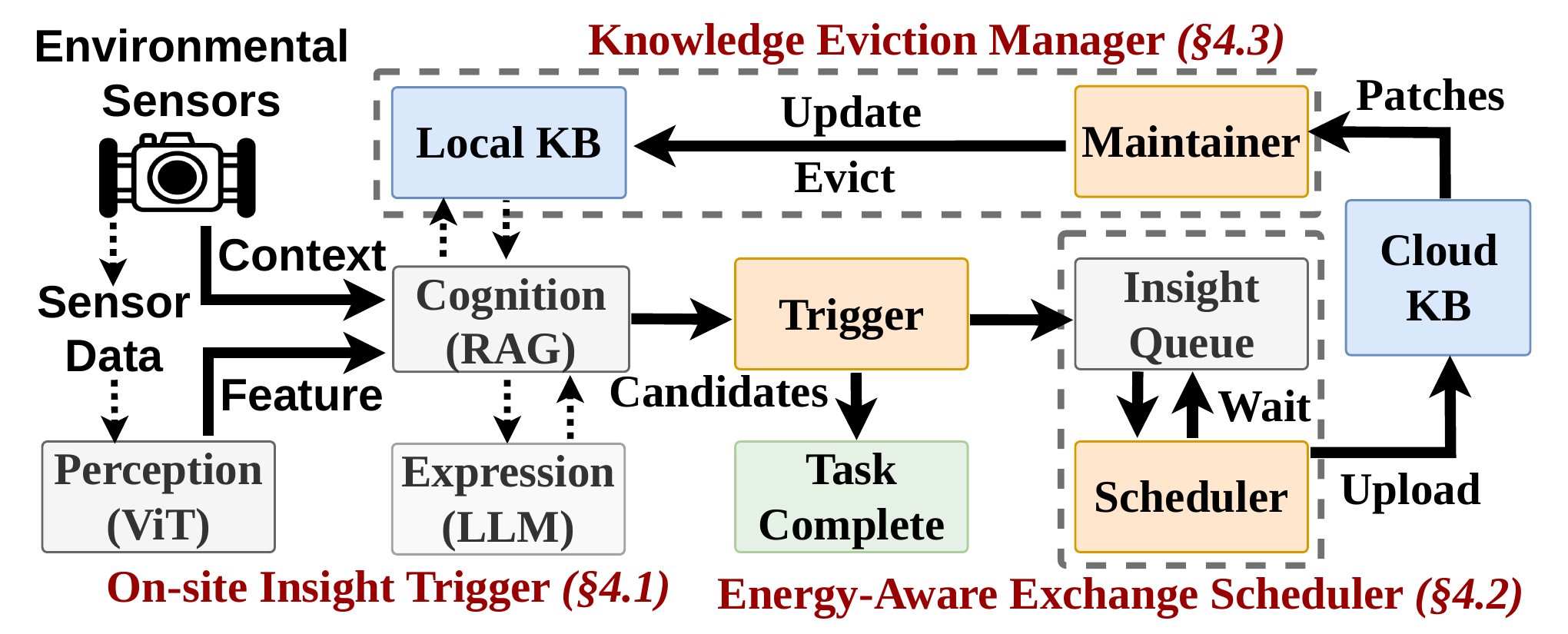}
    \vspace{-0.5cm}
    \caption{Overview of KADEX.}
    \label{fig:Workflow}
\end{figure}

As illustrated in Figure~\ref{fig:Workflow} and Algorithm~\ref{alg:kadex_lifecycle}, KADEX addresses these through three synchronized modules. The On-site Insight Trigger (§\ref{subsection:Trigger}) functions as a cognitive gatekeeper. It utilizes structural entropy to filter routine data and queue only high-value insights. The Energy-Aware Exchange Scheduler (§\ref{subsection:Scheduler}) optimizes satellite uplink usage. This module dynamically weighs informational urgency against real-time battery states. Finally, the Knowledge Eviction Manager (§\ref{subsection:Manager}) enforces storage constraints. It applies a community-level policy to retire outdated semantic regions, ensuring the local knowledge base remains relevant and coherent.

\begin{algorithm}[t]
\caption{KADEX High-Level Operational Lifecycle}
\label{alg:kadex_lifecycle}
\small
\begin{algorithmic}[1]
\Require Video stream $\{I_t\}$, L-SKG $\mathcal{G}_{\text{local}}$, Battery state $B(t)$
\Ensure Continuously adapted knowledge $\mathcal{G}_{\text{local}}$ and insight uploads

\State $\mathcal{Q}(t) \gets \emptyset$; \quad \text{SystemActive} $\gets$ \textbf{true};

\While{\text{SystemActive}}
    \State $I_t \gets$ CaptureFrame(); \quad $B(t) \gets$ UpdateBatteryState();

    \State \textcolor{commentcolor}{// Phase 1: On-site Insight Trigger (Cognitive Gatekeeper)}
    \State $G_{\text{inf}} \gets \text{PerceiveAndReason}(I_t, \mathcal{G}_{\text{local}})$;
    \If{$\text{IsHighEntropy}(G_{\text{inf}}, \tau_{\text{trigger}})$}
        \State \textcolor{commentcolor}{// Capture ambiguous frames for expert review}
        \State $\mathcal{Q}(t).\text{enqueue}(\text{PackInsight}(I_t))$;
    \Else
        \State \text{ExecuteLocalTask}($G_{\text{inf}}$); \quad \textcolor{commentcolor}{// Routine inference}
    \EndIf

    \State \textcolor{commentcolor}{// Phase 2: Energy-Aware Exchange Scheduler}
    \State $E_{\text{budget}} \gets \text{ComputeSafeBudget}(B(t), B_{\text{safe}})$;
    \If{$E_{\text{budget}} > 0$}
        \State Sort $\mathcal{Q}(t)$ by priority (Entropy descending);
        \State \text{TransmitInsights}(\text{Uplink}, $\mathcal{Q}(t)$, $E_{\text{budget}}$);
    \EndIf

    \State \textcolor{commentcolor}{// Phase 3: Community-Level Knowledge Manager}
    \If{received patches $\{\text{Patch}(c)\}$ from Cloud}
        \State $\mathcal{G}_{\text{local}} \gets \text{IntegratePatches}(\mathcal{G}_{\text{local}}, \{\text{Patch}(c)\})$;
        \State \textcolor{commentcolor}{// Maintain structural coherence under storage limits}
        \State \text{EnforceEvictionPolicy}($\mathcal{G}_{\text{local}}$, \text{Cap});
    \EndIf

    \State $t \gets t + 1$;
\EndWhile

\end{algorithmic}
\end{algorithm}

\subsection{On-site Insight Trigger}
\label{subsection:Trigger}

The On-site Insight Trigger determines whether a new frame reflects routine conditions or reveals patterns that the Local Structured Knowledge Graph (L-SKG) cannot explain. Thresholds based only on visual similarity are unreliable because they ignore ecological structure and explicit logical constraints. To address this limitation, we measure the semantic fit between the observation and the L-SKG using Structural Graph Entropy, which reflects whether the observation supports a single coherent interpretation or forces the system into uncertainty or contradiction.

The L-SKG is a compact graph stored on the edge device, defined as $G=(V,E)$. The node set $V$ contains entity nodes $V_{ent}$, attribute nodes $V_{att}$, and context nodes $V_{ctx}$. Entity nodes represent species such as chinook or sockeye. Attribute nodes encode observable morphological traits such as dorsal spots or a hooked lower jaw. Context nodes describe site and environmental factors such as site identity, GPS region, water temperature, seasonal stage, or diel lighting. The edge set contains only two semantic types: (1) supporting edges, which link attributes to entities with positive weights ($w(a,e)>0$), and (2) conflict edges, which specify mutually exclusive relationships grounded in expert knowledge.

When the perception module (e.g. CLIP~\cite{radford2021learning}) processes an image $I$, it outputs feature embeddings $V_f=\{f_1,\ldots,f_m\}$. Each feature activates a subset of attribute nodes, yielding the set $A_{feat}$. Sensor readings activate a set of context nodes $A_{ctx}$. The combined activation set is $A=A_{feat}\cup A_{ctx}$. From this set, the trigger constructs an \textbf{inference subgraph} $G_{i}$ that focuses only on knowledge relevant to the current frame. It first identifies the neighboring entity nodes $V_{ent}^*=N(A)\cap V_{ent}$. The node set and edge set of the inference subgraph are defined as follows:
$$
G_i=(V_i,E_i),\, V_i=A\cup V_{ent}^*,\, E_i=\{(u,v){\in}E:u,v{\in}V_i\}.
$$
This construction isolates the specific portion of the knowledge graph needed to interpret the current observation and excludes species or contextual knowledge unrelated to the active attributes and conditions.

Within this subgraph, the trigger computes a \textbf{semantic support score} for each candidate entity. It first checks for explicit contradictions: if any activated attribute $a\in A_{feat}$ connects to an entity $e$ through a conflict edge, then $e$ is incompatible with the observation and is assigned $S(e)=0$. For all remaining entities, the support score aggregates evidence from supporting edges:
\begin{equation}
S(e)=\sum_{a\in A_{feat}} w(a,e).
\end{equation}
Attributes without supporting edges contribute zero. This score increases when the observed traits match the species’ typical descriptors in the L-SKG and drops to zero when a hard conflict is detected.

Only entities with positive support form the valid set $E_{valid}=\{e\in V_{ent}^* \mid S(e)>0\}$. The trigger then normalizes support scores into a probability distribution:
\begin{equation}
P(e\mid V_f,C_x)= \frac{\exp(S(e))}{\sum_{e'\in E_{valid}}\exp(S(e'))},
\end{equation}
and assigns zero probability to all excluded entities. This distribution captures both the strength of positive evidence and the effect of explicit constraints that remove otherwise plausible classes.

The \textbf{Structural Graph Entropy} is the Shannon entropy of this distribution:
\begin{equation}
H(G_{i})=-\sum_{e\in E_{valid}} P(e\mid V_f,C_x)\log P(e\mid V_f,C_x).
\end{equation}
Entropy is low when one entity dominates, indicating that the L-SKG provides a stable, coherent explanation. Entropy rises when several explanations compete or when conflict edges eliminate what would otherwise be the leading entity. Thus, the entropy reflects both ambiguity and structural tension without requiring a separate conflict penalty.

The trigger compares the entropy to a threshold $\tau_{trigger}$. Frames with entropy below the threshold are routine and update only local statistics. Frames with entropy above the threshold are marked as high-value because they indicate missing knowledge or a violation of existing structure.

\subsection{Energy-Aware Exchange Scheduler}
\label{subsection:Scheduler}

The existing system relied on fixed rules for uplink control, i.e., preset 2-hour upload windows per day. These rules fail when the battery level and the rate of insight generation fluctuate. During long low-light periods, a fixed schedule may continue transmitting and drain the battery below a safe margin. Conversely, during bursts of high-entropy insights, a fixed cap cannot prioritize the most critical accumulated information. To address this, we model uplink scheduling as an energy-aware optimization problem and derive a closed-loop suitable for edge deployment.

As described in Sec~\ref{subsection:Trigger}, the On-site Insight Trigger filters raw frames and generates an Insight Packet only when the structural graph entropy \(H(G_{inf})\) is high. At each time slot \(t\), the device maintains a queue \(\mathcal{Q}(t)\) of such packets. Unlike simple FIFO buffers, the scheduler must prioritize observations that offer the highest information gain. Therefore, we sort \(\mathcal{Q}(t)\) by entropy \(H(p)\) in descending order. Each packet \(p \in \mathcal{Q}(t)\) stores its entropy, a fused feature vector \(z_p\), and a timestamp. With an estimate of the energy per bit, the cost of uploading a packet is approximated by a constant \(e_{\text{pkt}}\).

Let $B(t)$ denote the battery state and $B_{\text{safe}}$ a safety threshold that preserves energy for essential sensing and local inference. The scheduler assigns a dynamic communication budget:
\begin{equation}
E_{\text{budget}}(t) = \max\{0,\, B(t)-B_{\text{safe}}\},
\end{equation}
which grows when the battery is healthy and shrinks to zero during extended low-light periods, automatically pausing uplink activity.

Given this budget, the scheduler solves a per-slot optimization problem. Let the sorted insight queue at time $t$ be
\[
\mathcal{Q}(t)=\{p_1(t), p_2(t), \ldots, p_{n(t)}(t)\},
\]
where $H(p_1) \ge H(p_2) \ge \ldots \ge H(p_{n(t)})$. The decision variable
$
k(t) \in \{0, 1, \ldots, n(t)\}
$
specifies how many packets to upload from the head of the priority queue. The objective is to transmit the maximum total entropy allowed by the current energy constraint:
\begin{equation}
\max_{k(t)} \sum_{j=1}^{k(t)} H(p_j(t))
\quad
\text{s.t.} \quad
k(t)\, e_{\text{pkt}} \le E_{\text{budget}}(t).
\label{eq:energy_opt}
\end{equation}

Since the packets are already sorted by value (entropy), the optimal strategy is simply to transmit as many top-priority packets as the budget allows:
\begin{equation}
k^\star(t)
=
\min\!\left(n(t),\;
\left\lfloor \frac{E_{\text{budget}}(t)}{e_{\text{pkt}}} \right\rfloor\right).
\label{eq:closed_form_k}
\end{equation}

This prioritization keeps routine monitoring active under low energy. It also moves the most useful novel cases to the front of the upload queue when communication is available.

Once uploaded, high-entropy packets drive the expert-in-the-loop workflow. This is not full stream annotation. Experts review only cases that remain unclear after cloud retrieval. If the cloud master graph can explain the packet, it returns a knowledge patch to the edge. Otherwise, an expert gives a short text or audio note. A knowledge graph-based RAG pipeline~\cite{edge2024local} converts the note into structured nodes and edges. The cloud master graph is updated first. The edge receives only the relevant patch.

\subsection{Knowledge Eviction Manager}
\label{subsection:Manager}

After uplink transmission, the cloud processes the uploaded Insight Packets and applies them to the cloud master graph $\mathcal{G}_{\text{global}}$. Based on this cloud view, the cloud generates device-specific knowledge patches. These patches are naturally organized by semantic communities. A community may correspond to a species (e.g., \textit{chinook}), a species and season combination (e.g., \textit{chinook–spawning}), a species and sex category, or even finer-grained variants (e.g., \textit{jack chinook}). Each patch contains the subgraph of that community together with its local updates, including new nodes, edges, and associated weights. When delivered to the device, these patches incrementally update the L-SKG.

Conventional LRU policies operate at the level of individual items, which here correspond to individual nodes or edges. Such fine-grained eviction is unsuitable for our inference procedure in Section~3.2. The inference subgraph $G_{\text{inf}}$ is constructed by expanding around a set of activated nodes and collecting their neighbors. Evicting isolated nodes or edges breaks the neighborhood structure inside a semantic community, leaving incomplete local contexts and inconsistent constraint sets. To avoid this fragmentation, we use a \emph{community-level} LRU policy. All loading, updating, and eviction occur at the granularity of semantic communities, preserving the structural coherence needed for local inference.

The device stores its L-SKG as a union of community subgraphs. Let $\mathcal{C}_{\text{local}}$ be the current set of cached communities, and let each $c \in \mathcal{C}_{\text{local}}$ have an associated subgraph $\mathcal{G}_c^{\text{local}}$ and a timestamp $\text{LastUsed}(c)$. This timestamp is refreshed whenever the community is accessed during inference or updated by an incoming patch. The local storage is bounded by a capacity $\text{Cap}$, measured in nodes, edges, or total bytes. When new patches arrive, they may cause the L-SKG to exceed this capacity. In that case, the system performs LRU eviction at the community level: communities with the earliest $\text{LastUsed}$ values are removed entirely until the capacity constraint is satisfied.

This community-level LRU strategy ensures that the device maintains coherent semantic regions of the L-SKG. By evicting entire communities rather than isolated items, the method avoids breaking neighborhood structure and keeps the local graph aligned with the inference process in Sec~\ref{subsection:Trigger}. Under limited storage, the device automatically retains the communities that have been most relevant to recent observations, providing a compact and stable prior for subsequent entropy-based inference.

\section{Case Study and Implementation Plan}
Our implementation plan follows a staged pathway that couples system hardening with knowledge curation, moving from an expert-built cloud knowledge base to an expert-in-the-loop field workflow, and finally to expert-bounded autonomous operation. We start from a practical observation in this domain: the limiting factor for small on-device models is often not feature extraction, but the absence of explicit expert logic. KADEX therefore treats knowledge as a first-class artifact that is constructed, audited, and updated through lightweight patches, while keeping on-site computation and communication bounded.

\paragraph{Phase 1: Cloud knowledge base bootstrapping from expert narratives.}
We first bootstrap a master knowledge base in the cloud from detailed species explanations provided by domain experts. These materials are recorded as structured audio narratives. They cover diagnostic traits, exclusionary rules, life-stage transitions, and site/season-specific priors. We then transcribe them into text and convert them into a structured knowledge graph schema. This process turns rich expertise into explicit entities and constraints. Examples include trait--species relations, negative rules, and spatiotemporal applicability. The result is an auditable starting point that does not depend on data-intensive retraining.

\paragraph{Phase 2: Expert-in-the-loop deployment and iterative knowledge patching.}
We then deploy KADEX at partner-designated sites under real constraints. These include duty-cycled satellite connectivity and solar/battery energy budgets. During this stage, KADEX runs with an expert-in-the-loop workflow. The \textbf{On-site Insight Trigger} surfaces only high-uncertainty or novel cases as compact insight packets. Local domain experts review these packets and correct the reasoning when needed. They submit updates as text or short audio notes. These corrections become knowledge patches. The patches update the cloud master graph and are selectively pushed back to edge devices. This allows the fleet to adapt without uploading raw video or retraining model weights. Experts review only unresolved high-entropy packets, not the full stream. One active site may need tens to low hundreds of expert hours in early Phase~2, then periodic auditing.

\paragraph{Phase 3: Expert-bounded autonomy with preserved digital heritage.}
Once the agent's behavior stabilizes, KADEX can operate autonomously for low-entropy routine observations. This autonomy remains bounded by expert authority. It conserves scarce expert time for high-entropy observations, unknown species or life stages, site-specific ecological change, sensor failures, and management-relevant decisions. The knowledge base then becomes both durable \emph{digital heritage} and an interface for expert validation. Expert logic is preserved in an explicit form. New patches still require expert validation, revision, or abstention when the system lacks a reliable explanation.

\paragraph{Case study and resources.} We will use existing multi-site historical data to replay streams and test the full KADEX lifecycle, then transition to the partner-provided field pilot for iterative patching and hardening. The project already has core Phase~1 infrastructure, including edge compute nodes (Jetson-class devices\footnote{\url{https://www.nvidia.com/en-us/autonomous-machines/embedded-systems/}}), solar harvesting and battery storage, and historical observations. The main Phase~2 requirement is sustained access to domain experts who validate surfaced insight packets and issue structured knowledge updates through text or short audio notes.

\section{Evaluation Criteria}
To ensure the proposed solution meets the rigorous demands of scientific monitoring and off-grid operation, we establish a multi-dimensional evaluation framework. This protocol encompasses ecological precision, operational sustainability, and communication efficiency, grounded in the constraints identified during our preliminary pilot analysis.

\subsection{Taxonomic Precision and Reliability}
The primary success metric is the system's ability to replicate expert-level classification in the wild. We evaluate performance using F1-score~\cite{goodfellow2016deep} and mean Average Precision (mAP) on expert-annotated longitudinal datasets. To validate the efficacy of our approach, we benchmark against three categories of baselines. First, we compare results with standard real-time object detectors, specifically YOLO~\cite{kay2022caltech} and RT-DETR~\cite{zhao2024detrs}. Second, we evaluate performance against the domain-specific STSVT system~\cite{xu2024salina}. Third, we assess our method against advanced multimodal foundation models, including VLM (e.g. Qwen3-VL~\cite{bai2025qwen3}) and recent fusion approaches~\cite{xu2025exploring}. These comparisons focus on fine-grained classification accuracy for visually similar species and the system's ability to reject non-target ``confuser'' species.
Initial experiments suggest that integrating expert logic helps mitigate visual hallucinations common in VLMs. This approach shows promise for distinguishing visually similar species.

\subsection{Survival Under Intermittent Energy} For off-grid viability, we evaluate system resilience using loss-of-load probability (LOLP)~\cite{allan2013reliability}. This standard metric quantifies the likelihood of the Battery State of Charge (SOC) dropping below the critical safety threshold (30\%) over a long-term deployment. We compare our adaptive policy against the fixed duty-cycle baseline~\cite{xu2024salina}, which enforces a static two-hour daily transmission window. The evaluation measures the reduction in LOLP under constrained energy profiles, such as prolonged overcast periods. Our target is to achieve an LOLP of zero (indicating continuous availability) in scenarios where the baseline experiences forced shutdowns.
Initial simulations using historical solar irradiance data indicate that our energy-aware scheduler successfully maintains the battery state of charge above the safety margin during extended low-light intervals.

\section{Expected Results and Impacts}
This project aims to make AI monitoring practical in remote ecosystems, while preserving expert knowledge in a form that can be reused and updated over time.

\paragraph{Expected technical outcomes.}
KADEX replaces repeated cloud retraining with lightweight knowledge updates. We expect this to reduce energy use and data transfer compared to cloud-centric model adaptation, which helps maintain stable operation under solar and battery limits. We also expect more reliable and accurate fine-grained identification in the field, as each decision is constrained by an explicit expert knowledge graph rather than being hidden in model weights. Evaluation will follow a staged process: offline replay on multi-site historical data, then expert-in-the-loop validation in a partner-designated pilot.

\paragraph{Expected societal and scientific impacts.}
KADEX promotes knowledge sustainability by converting tacit, local expertise into explicit and reusable rules and graph structures. This reduces knowledge loss when experts retire or teams change. KADEX also makes system updates easier for practitioners. Instead of retraining models, they can improve the system by adding or revising rules, priors, and context descriptions. Finally, KADEX lowers the barrier to long-term monitoring in under-resourced settings by reducing dependence on continuous connectivity and large data uploads. The same approach can extend beyond salmon to other biodiversity tasks where lookalike species and changing conditions make vision-only models unreliable. Because expert time is scarce, KADEX shifts effort from full-stream annotation to graph validation, anomaly review, and management decisions. As physical AI matures~\cite{nvidia2026cosmos}, robots may collect richer field evidence. Experts will still be needed for knowledge-base design, sampling, site context, and cases beyond sensor and physical AI coverage.

\vspace{-0.13cm}

\section{Practical Challenges}
Deploying knowledge-adaptive agents in infrastructure-constrained wild environments remains uncertain. In particular, the system's initial reasoning depends on bootstrapped expert knowledge, and ``unknown'' species (unknown unknowns) challenge closed-world graphs. KADEX therefore treats high-entropy events as review triggers. If a case is unknown only to the L-SKG, the cloud master graph sends a knowledge patch back to the edge. If the case is also unknown to the cloud master graph, experts review a compact packet and update the graph with text or audio notes. If experts cannot assign a reliable species label, the system stores the case as an anomaly with time, location, and visual evidence. It does not force a classification. During long outages, low-entropy cases still run locally, while high-entropy packets wait until energy and connectivity allow upload. This workflow lets experts patch the graph over time and turns field intuition into reusable digital heritage.

In addition, hardware constraints impose physical trade-offs. Finite edge storage requires eviction of inactive knowledge communities, which may remove local context for rare or seasonal species. The cloud master graph limits this risk. It grows only when verified and reusable patches are available. Unresolved anomalies are not added as low-confidence species nodes, which helps limit graph noise. When an evicted species reappears, the edge uploads a high-entropy packet and receives the relevant cloud subgraph. Before patching, the system may abstain or produce a false negative.

\newpage
\appendix
\section*{Project Team Description}


This research brings together an interdisciplinary team spanning AI, networked systems, and fisheries science. The Simon Fraser University (SFU) team leads the technical development of the KADEX architecture, focusing on edge computing, multimodal RAG, and energy-aware control. In parallel, our domain partners (fisheries biologists and Indigenous stewardship practitioners) co-define monitoring goals and constraints, contribute localized ecological knowledge, and validate outputs for field deployment.

\medskip

\noindent\textbf{Jiaxing Li}\\
\textit{Ph.D. Candidate, Simon Fraser University, Canada}\\
Jiaxing Li is currently a Ph.D. candidate in the School of Computing Science at Simon Fraser University, British Columbia, Canada, under the supervision of Prof. Jiangchuan Liu. He received his B.Sc. degree in Computing Science from Simon Fraser University in 2023. His research interests include large language model applications, cloud edge computing, and distributed machine learning systems.

\medskip

\noindent\textbf{Hao Fang}\\
\textit{Ph.D. Student, Simon Fraser University, Canada}\\
Hao Fang received his B.Sc. (Hons.) degree with distinction in Computing Science from Simon Fraser University, BC, Canada, in 2022. He is currently a Ph.D. student in Computing Science at Simon Fraser University. His research areas include satellite communications and networking, particularly with multimedia systems.


\medskip

\noindent\textbf{Chi Xu}\\
\textit{Ph.D. Candidate, Simon Fraser University, Canada}\\
Chi Xu received the B.Sc. degree in software engineering from Xidian University, Xi'an, China, and the M.Sc. degree in computing science from Simon Fraser University. He is currently pursuing his Ph.D. degree in computing science at Simon Fraser University, Canada. His research focuses on multimodal data sensing, management, and analytics, spanning the fields of Artificial Internet of Things (AIoT) and networked systems.

\medskip

\noindent\textbf{Miao Zhang}\\
\textit{Postdoctoral Researcher, Simon Fraser University, Canada}\\
Miao Zhang received her Ph.D., M.Eng., and B.Eng. degrees in Computer Science and Technology from Simon Fraser University, Tsinghua University, and Sichuan University, respectively. She is currently a postdoctoral researcher at Simon Fraser University. Her research interests include cloud computing, edge computing, and multimedia systems.

\medskip

\noindent\textbf{Dr. Jiangchuan Liu}\\
\textit{Professor, Simon Fraser University, Canada}\\
Dr. Jiangchuan Liu is a Professor in the School of Computing Science, Simon Fraser University, British Columbia, Canada. He is a Fellow of the Royal Society of Canada, a Fellow of the Canadian Academy of Engineering, an IEEE Fellow, and an NSERC E.W.R. Steacie Memorial Fellow. His research interests include intelligent multimedia computing and networking, cloud and edge computing, and wireless mobile and space networking. He is a co-recipient of the IEEE INFOCOM Test of Time Paper Award (2015), IEEE ICDCS Distinguished Paper Award (2024), ACM SIGMM TOMCCAP Nicolas D. Georganas Best Paper Award (2013), and ACM Multimedia Best Paper Award (2012). He has served on the editorial boards of IEEE/ACM TON, IEEE TNSE, TMM, TBD, COMST, and IOTJ, and was TPC Chair of IEEE INFOCOM 2021 and General Chair of INFOCOM 2024.

\medskip

\noindent\textbf{Dr. William I. Atlas}\\
\textit{Salmon Watershed Scientist, Wild Salmon Center, USA}\\
Dr. William I. Atlas serves as a Salmon Watershed Scientist at the Wild Salmon Center (WSC). Before his tenure at WSC began in 2020, Dr. Atlas dedicated over a decade to partnering with Central Coast First Nations (CCFN), where he played a pivotal role in co-designing community-led salmon research initiatives during both his graduate studies and professional career. Notably, during his postdoctoral fellowship at the Pacific Salmon Foundation (PSF), he orchestrated the creation of the Central Coast Monitoring Framework. Currently, Dr. Atlas represents the Central Coast region on the Northern Panel of the Pacific Salmon Commission (PSC). As an active member of the PSC's First Nations Caucus, he provides essential strategic guidance for CCFN engagement and ensures community leadership is kept informed regarding bilateral management negotiations.

\medskip

\noindent\textbf{Dr. Katrina M. Connors}\\
\textit{Senior Director of Salmon Programs, Pacific Salmon Foundation, Canada}\\
Dr. Katrina M. Connors currently serves as the Senior Director of Salmon Programs at the Pacific Salmon Foundation (PSF), where she provides strategic oversight for marine and freshwater science, data infrastructure, community outreach, and grant administration. With nearly two decades of professional experience, Dr. Connors has directed extensive collaborative research aimed at the conservation and management of Pacific salmon. As the founding director of the PSF's Salmon Watersheds Program, she spearheaded the creation of the Pacific Salmon Explorer, an essential open-access tool for visualizing salmon populations and habitat data throughout British Columbia. She also guided the publication of Canada's inaugural State of Salmon Report, establishing a data-driven baseline for assessing salmon trends across BC and the Yukon. Additionally, Dr. Connors represents national interests as a Canadian Commissioner for the Pacific Salmon Commission (PSC).


\medskip

\noindent\textbf{Mark A. Spoljaric}\\
\textit{Program Biologist, Haida Fisheries Program, Canada}\\
Mark A. Spoljaric is a seasoned fisheries biologist with over ten years of experience developing and managing salmon research and conservation initiatives on Haida Gwaii. In his role as a Program Biologist for the Haida Fisheries Program, he manages stock escapement surveys and coordinates project planning for multiple salmon-bearing watersheds. His technical fieldwork includes habitat assessments in remote coastal environments and the enumeration of returning salmon and out-migrating juvenile fish. Beyond his research duties, Mr. Spoljaric contributes to public engagement efforts focused on salmon conservation. In addition, he serves on the Northern Boundary Technical Committee of the Pacific Salmon Commission (PSC) and is an active member of the PSC's First Nations Caucus.



\section*{Ethics Statement}
This research follows a ``Responsibility-by-Design'' framework that prioritizes ethical standards, local rights, and Indigenous Data Sovereignty. Local partners retain ownership of raw data and control how sensitive information is shared, which helps prevent exploitation and preserve local authority. For privacy, KADEX processes data on the edge, filters images of human activities locally, and transmits only ecological data. For accountability, KADEX uses an explicit knowledge graph rather than an opaque ``black-box'' model, allowing experts to audit decision logic and correct biased or unsafe patches. Human oversight is treated as meaningful authority, not symbolic approval. Local experts and Indigenous partners must be able to inspect evidence paths, reject or freeze unsafe knowledge patches, and define which decisions require human or community review.

\section*{Acknowledgments}
This work was supported in part by an NSERC Discovery Grant, a British Columbia Salmon Recovery and Innovation Fund (BCSRIF 2022 401), a Mitacs Accelerate Cluster Grant, and support from experiment.com. We thank the Heiltsuk, Haida, Kitasoo Xai'xais, Taku River Tlingit, and Gitga'at First Nations for their trust, guidance, and collaboration. We also thank the Skeena Fishery Commission and the Gitanyow Fisheries Authority for their partnership in field monitoring and salmon stewardship. The corresponding author for this work is Jiangchuan Liu.

\bibliographystyle{named}
\bibliography{ijcai26}

@misc{musmanni2023protecting,
      title={Protecting Wildlife in a Changing Climate: Four Powerful Adaptation Strategies},
      author={D{\'i}az Musmanni, Gabriela},
      year={2023},
      howpublished = {\url{https://gca.org/protecting-wildlife-in-a-changing-climate-four-powerful-adaptation-strategies/}}
    }

@techreport{wwf2024living,
      title       = {Living Planet Report 2024: A System in Peril},
      author      = {{World Wide Fund for Nature}},
      institution = {WWF},
      year        = {2024},
      isbn        = {978-2-88085-319-8},
      type        = {Report},
      url         = {https://wwflpr.awsassets.panda.org/downloads/2024-living-planet-report-a-system-in-peril.pdf}
    }

@misc{naturetech2025ground,
      title={Exploring Ground-Truth Nature Tech \& the Future of Biodiversity Monitoring},
      author={{Nature Tech Collective}},
      year={2025},
      howpublished = {\url{https://www.naturetechcollective.org/stories/ground-truth-biodiversity-wildlife-monitoring}}
    }

@article{norouzzadeh2018automatically,
  title={Automatically identifying, counting, and describing wild animals in camera-trap images with deep learning},
  author={Norouzzadeh, Mohammad Sadegh and Nguyen, Anh and Kosmala, Margaret and Swanson, Alexandra and Palmer, Meredith S and Packer, Craig and Clune, Jeff},
  journal={Proceedings of the National Academy of Sciences (PNAS)},
  year={2018}
}

@article{hill2025starlink,
  title={Starlink speeds rise, but still fall (mostly) short of FCC standards},
  author={Hill, Kelly},
  journal={RCR Wireless News},
  year={2025}
}

@inproceedings{paul2024simple,
  title={A Simple Interpretable Transformer for Fine-Grained Image Classification and Analysis},
  author={Paul, Dipanjyoti and Chowdhury, Arpita and Xiong, Xinqi and Chang, Feng-Ju and Carlyn, David Edward and Stevens, Samuel and Provost, Kaiya and Karpatne, Anuj and Carstens, Bryan and Rubenstein, Daniel I and others},
  booktitle={International Conference on Learning Representations (ICLR)},
  year={2024}
}

@article{axford2024collectively,
  title={Collectively advancing deep learning for animal detection in drone imagery: Successes, challenges, and research gaps},
  author={Axford, Daniel and Sohel, Ferdous and Vanderklift, Mathew A and Hodgson, Amanda J},
  journal={Ecological Informatics},
  year={2024}
}

@inproceedings{xu2024salina,
  title={SALINA: Towards Sustainable Live Sonar Analytics in Wild Ecosystems},
  author={Xu, Chi and Qian, Rongsheng and Fang, Hao and Ma, Xiaoqiang and Atlas, William I. and Liu, Jiangchuan and Spoljaric, Mark A.},
  booktitle={ACM Conference on Embedded Networked Sensor Systems (SenSys)},
  year={2024}
}

@inproceedings{xu2025exploring,
  title={Exploring Multimodal Foundation AI and Expert-in-the-Loop for Sustainable Management of Wild Salmon Fisheries in Indigenous Rivers},
  author={Xu, Chi and Jin, Yili and Ma, Sami and Qian, Rongsheng and Fang, Hao and Liu, Jiangchuan and Liu, Xue and Ngai, Edith CH and Atlas, William I and Connors, Katrina M and others},
  booktitle={International Joint Conference on Artificial Intelligence (IJCAI)},
  year={2025}
}

@inproceedings{xu2026fused,
  title={FUSED: Toward Federated Multimodal Retrieval across Sovereign Data Domains},
  author={Xu, Chi and Li, Jiaxing and Jin, Mengdi and Atlas, William I. and Spoljaric, Mark A. and Ngai, Edith C. H. and Liu, Jiangchuan},
  booktitle={The ACM Web Conference (WWW)},
  year={2026}
}

@article{ma2024leo,
  title={LEO satellite network access in the wild: Potentials, experiences, and challenges},
  author={Ma, Sami and Chou, Yi Ching and Zhang, Miao and Fang, Hao and Zhao, Haoyuan and Liu, Jiangchuan and Atlas, William I},
  journal={IEEE Network},
  year={2024}
}

@article{folkman2025data,
  title={A data-centric framework for combating domain shift in underwater object detection with image enhancement},
  author={Folkman, Lukas and Pitt, Kylie A and Stantic, Bela},
  journal={Applied Intelligence},
  year={2025}
}

@article{yang2024does,
  title={Does negative sampling matter? A review with insights into its theory and applications},
  author={Yang, Zhen and Ding, Ming and Huang, Tinglin and Cen, Yukuo and Song, Junshuai and Xu, Bin and Dong, Yuxiao and Tang, Jie},
  journal={IEEE Transactions on Pattern Analysis and Machine Intelligence (TPAMI)},
  year={2024}
}

@article{bai2025qwen3,
  title={Qwen3-VL Technical Report},
  author={Bai, Shuai and Cai, Yuxuan and Chen, Ruizhe and Chen, Keqin and Chen, Xionghui and Cheng, Zesen and Deng, Lianghao and Ding, Wei and Gao, Chang and others},
  journal={arXiv preprint arXiv:2511.21631},
  year={2025}
}

@inproceedings{lewis2020rag,
  title={Retrieval-Augmented Generation for Knowledge-Intensive NLP Tasks},
  author={Lewis, Patrick and Perez, Ethan and Piktus, Aleksandra and Petroni, Fabio and Karpukhin, Vladimir and Goyal, Naman and Kuttler, Heinrich and Lewis, Mike and Yih, Wen-tau and Rockt{\"a}schel, Tim and Riedel, Sebastian and Kiela, Douwe},
  booktitle={International Conference on Neural Information Processing Systems (NeurIPS)},
  year={2020}
}

@inproceedings{li2024structrag,
  title={StructRAG: Boosting Knowledge-Intensive Reasoning of Large Language Models via Inference-Time Hybrid Information Structurization},
  author={Li, Zeyu and Guo, Qipeng and Yao, Yunzhi and others},
  booktitle={International Conference on Learning Representations (ICLR)},
  year={2024}
}

@article{yu2024visrag,
  title={VisRAG: Vision-Based Retrieval-Augmented Generation on Multi-Modality Documents},
  author={Yu, Shi and Tang, Chaoyue and Xu, Bokai and others},
  journal={arXiv preprint arXiv:2410.10594},
  year={2024}
}

@book{allan2013reliability,
      title={Reliability evaluation of power systems},
      author={Allan, Ronald N and others},
      year={2013},
      publisher={Springer Science \& Business Media}
    }

@article{edge2024local,
  title={From local to global: A graph rag approach to query-focused summarization},
  author={Edge, Darren and Trinh, Ha and Cheng, Newman and Bradley, Joshua and Chao, Alex and Mody, Apurva and Truitt, Steven and Metropolitansky, Dasha and Ness, Robert Osazuwa and Larson, Jonathan},
  journal={arXiv preprint arXiv:2404.16130},
  year={2024}
}

@article{waples2008evolutionary,
  title={Evolutionary history of Pacific salmon in dynamic environments},
  author={Waples, Robin S and Pess, George R and Beechie, Tim},
  journal={Evolutionary Applications},
  year={2008}
}

@article{di2016multi,
  title={Multi-year persistence of the 2014/15 North Pacific marine heatwave},
  author={Di Lorenzo, Emanuele and Mantua, Nathan},
  journal={Nature Climate Change},
  year={2016}
}

@article{frolicher2018emerging,
  title={Emerging risks from marine heat waves},
  author={Fr{\"o}licher, Thomas L and Laufk{\"o}tter, Charlotte},
  journal={Nature communications},
  year={2018}
}

@article{kilduff2015changing,
  title={Changing central Pacific El Ni{\~n}os reduce stability of North American salmon survival rates},
  author={Kilduff, D Patrick and Di Lorenzo, Emanuele and Botsford, Louis W and Teo, Steven LH},
  journal={Proceedings of the National Academy of Sciences},
  year={2015}
}

@article{dorner2018spatial,
  title={Spatial and temporal patterns of covariation in productivity of Chinook salmon populations of the northeastern Pacific Ocean},
  author={Dorner, Brigitte and Catalano, Matthew J and Peterman, Randall M},
  journal={Canadian Journal of Fisheries and Aquatic Sciences},
  year={2018}
}

@article{atlas2021indigenous,
  title={Indigenous systems of management for culturally and ecologically resilient Pacific salmon (Oncorhynchus spp.) fisheries},
  author={Atlas, William I and Ban, Natalie C and Moore, Jonathan W and Tuohy, Adrian M and Greening, Spencer and Reid, Andrea J and Morven, Nicole and White, Elroy and Housty, William G and Housty, Jess A and others},
  journal={BioScience},
  year={2021}
}

@article{schindler2015prediction,
  title={Prediction, precaution, and policy under global change},
  author={Schindler, Daniel E and Hilborn, Ray},
  journal={Science},
  year={2015}
}

@inproceedings{kay2022caltech,
  title={The Caltech Fish Counting dataset: a benchmark for multiple-object tracking and counting},
  author={Kay, Justin and Kulits, Peter and Stathatos, Suzanne and Deng, Siqi and Young, Erik and Beery, Sara and Van Horn, Grant and Perona, Pietro},
  booktitle={European Conference on Computer Vision},
  year={2022}
}

@article{atlas2023wild,
  title={Wild salmon enumeration and monitoring using deep learning empowered detection and tracking},
  author={Atlas, William I and Ma, Sami and Chou, Yi Ching and Connors, Katrina and Scurfield, Daniel and Nam, Brandon and Ma, Xiaoqiang and Cleveland, Mark and Doire, Janvier and Moore, Jonathan W and others},
  journal={Frontiers in Marine Science},
  year={2023}
}

@article{kay2024align,
  title={Align and Distill: Unifying and Improving Domain Adaptive Object Detection},
  author={Kay, Justin and Haucke, Timm and Stathatos, Suzanne and Deng, Siqi and Young, Erik and Perona, Pietro and Beery, Sara and Van Horn, Grant},
  journal={arXiv preprint arXiv:2403.12029},
  year={2024}
}

@article{wu2022survey,
  title={A survey of human-in-the-loop for machine learning},
  author={Wu, Xingjiao and Xiao, Luwei and Sun, Yixuan and Zhang, Junhang and Ma, Tianlong and He, Liang},
  journal={Future Generation Computer Systems},
  year={2022}
}

@book{goodfellow2016deep,
  title={Deep learning},
  author={Goodfellow, Ian},
  year={2016},
  publisher={MIT press}
}

@article{wang2024yolov10,
  title={Yolov10: Real-time end-to-end object detection},
  author={Wang, Ao and Chen, Hui and Liu, Lihao and Chen, Kai and Lin, Zijia and Han, Jungong and Ding, Guiguang},
  journal={arXiv preprint arXiv:2405.14458},
  year={2024}
}

@inproceedings{zhao2024detrs,
  title={Detrs beat yolos on real-time object detection},
  author={Zhao, Yian and Lv, Wenyu and Xu, Shangliang and Wei, Jinman and Wang, Guanzhong and Dang, Qingqing and Liu, Yi and Chen, Jie},
  booktitle={Proceedings of the IEEE/CVF Conference on Computer Vision and Pattern Recognition (CVPR)},
  year={2024}
}

@inproceedings{liu24fish,
  title={Benchmarking Fish Dataset and Evaluation Metric in Keypoint Detection
                  - Towards Precise Fish Morphological Assessment in Aquaculture Breeding},
  author={Weizhen Liu and
                  Jiayu Tan and
                  Guangyu Lan and
                  Ao Li and
                  Dongye Li and
                  Le Zhao and
                  Xiaohui Yuan and
                  Nanqing Dong},
  booktitle={International Joint Conference on Artificial Intelligence (IJCAI)},
  year={2024}
}

@inproceedings{gordon23rhinos,
  title={Find Rhinos without Finding Rhinos: Active Learning with Multimodal
                  Imagery of South African Rhino Habitats},
  author={Lucia Gordon and
                  Nikhil Behari and
                  Samuel Collier and
                  Elizabeth Bondi{-}Kelly and
                  Jackson A. Killian and
                  Catherine Ressijac and
                  Peter Boucher and
                  Andrew Davies and
                  Milind Tambe},
  booktitle={International Joint Conference on Artificial Intelligence (IJCAI)},
  year={2023}
}

@inproceedings{kshitiz23bird,
  title={Long-term Monitoring of Bird Flocks in the Wild},
  author={Kshitiz and
                  Sonu Shreshtha and
                  Ramy Mounir and
                  Mayank Vatsa and
                  Richa Singh and
                  Saket Anand and
                  Sudeep Sarkar and
                  Sevaram Mali Parihar},
  booktitle={International Joint Conference on Artificial Intelligence (IJCAI)},
  year={2023}
}

@inproceedings{radford2021learning,
  title={Learning transferable visual models from natural language supervision},
  author={Radford, Alec and Kim, Jong Wook and Hallacy, Chris and Ramesh, Aditya and Goh, Gabriel and Agarwal, Sandhini and Sastry, Girish and Askell, Amanda and Mishkin, Pamela and Clark, Jack and others},
  booktitle={International conference on machine learning},
  year={2021}
}

@misc{nvidia2026cosmos,
  author       = {{NVIDIA}},
  title        = {NVIDIA Cosmos: An Open Platform for Physical AI with World Foundation Models},
  year         = {2026},
  howpublished = {\url{https://www.nvidia.com/en-us/ai/cosmos/}}
}

@misc{un-2030agenda,
  author       = {{United Nations}},
  title        = {Transforming our World: The 2030 Agenda for Sustainable Development},
  year         = {2015},
  howpublished = {\url{https://sdgs.un.org/2030agenda}}
}

@misc{un-lnob,
  author       = {{United Nations}},
  title        = {Leave No One Behind},
  year         = {2021},
  howpublished = {\url{https://unsdg.un.org/2030-agenda/universal-values/leave-no-one-behind}}
}

\end{document}